\def\year{2021}\relax
\documentclass[letterpaper]{article} 
\usepackage{aaai21}  
\usepackage{times}  
\usepackage{helvet} 
\usepackage{courier}  
\usepackage[hyphens]{url}  
\usepackage{graphicx} 
\usepackage{natbib}  
\usepackage{caption} 
\usepackage{algorithm}
\usepackage[noend]{algpseudocode}
\usepackage{amsmath}
\usepackage{dsfont}
\usepackage[switch]{lineno}  %
\usepackage{color,soul}

\title{Towards interpretability of Mixtures of Hidden Markov Models}

\author {
        Negar Safinianaini,\textsuperscript{\rm 1}
        Henrik Boström \textsuperscript{\rm 1}
}

\affiliations {
    \\\textsuperscript{\rm 1} School of Electrical Engineering and Computer Science, KTH Royal Institute of Technology, Stockholm, Sweden \\
    negars@kth.se, bostromh@kth.se
}

\begin{document}
\maketitle


\begin{abstract}
Mixtures of Hidden Markov Models (MHMMs) are frequently used for clustering of sequential data. An important aspect of MHMMs, as of any clustering approach, is that they can be interpretable, allowing for novel insights to be gained from the data. However, without a proper way of measuring interpretability, the evaluation of novel contributions is difficult and it becomes practically impossible to devise techniques that directly optimize this property. In this work, an information-theoretic measure (entropy) is proposed for interpretability of MHMMs, and based on that, a novel approach to improve model interpretability is proposed, i.e., an entropy-regularized Expectation Maximization (EM) algorithm. The new approach aims for reducing the entropy of the Markov chains (involving state transition matrices) within an MHMM, i.e., assigning higher weights to common state transitions during clustering. It is argued that this entropy reduction, in general, leads to improved interpretability since the most influential and important state transitions of the clusters can be more easily identified. An empirical investigation shows that it is possible to improve the interpretability of MHMMs, as measured by entropy, without sacrificing (but rather improving) clustering performance and computational costs, as measured by the v-measure and number of EM iterations, respectively.

\end{abstract}


\section{Introduction}
Clustering is an important machine learning task, which concerns extracting information from data; naturally, the extracted information should be \textit{interpretable}, especially in decision-critical domains \cite{inter}. There is no widely adopted definition of interpretability within machine learning, but various definitions have been proposed \cite{inter-survey}. Here, we loosely define interpretability of a clustering method as its capability to reveal the most influential components (e.g. features) in forming the clusters \cite{inter-survey}. Clustering algorithms often provide limited support for allowing the user to understand the rationale for how clusters are formed, limiting their interpretability \cite{interclust}. In this work, we focus on interpretability of sequential data clustering methods which include feature-based, deep-learning, and model-based methods \cite{ts-survey,ts-nips,bishop}. Model-based methods are often favored for being probabilistic and transparent by composing fundamental constructs (model variables and their dependencies through conditional probability) that can explain the data \cite{model-based}. Motivated by these advantages, we concentrate on model-based approaches; more specifically on the Mixture of Hidden Markov Models (MHMM), designed for clustering of sequential data by \cite{MHMM}. MHMMs have been used for various tasks, such as time series clustering, music analysis, motion recognition, handwritten digit recognition, financial and medical analysis \cite{hmm-dtw,music,motion,momMHMM,finance,plos}. Various approaches have been proposed to improve upon using Expectation Maximization (EM), as introduced by \cite{dempster1977maximum}, which is the standard approach of inferring MHMMs \cite{bishop}. For example, in \cite{momMHMM} and \cite{spaMHMM}, the improvements are w.r.t. time complexity and sparsity of the model, respectively.

In medical applications, Hidden Markov Models (HMM) have been used to, e.g., capture the structure of disease patterns \cite{healthstate}, clinical changes in the progression of Alzheimer’s Disease \cite{AD}, and cell variation in cancer study \cite{hmmcopy} by modeling the variation patterns as state transitions of HMMs. The notion of interpretability is rarely mentioned in such studies and consequently not quantified. Still, the models are sometimes interpreted by medical experts. In some previous work, interpretability of HMMs is touched upon; in \cite{protein}, a biologically-motivated regularization is introduced in EM for HMMs, reducing the mean difference among the states, which provides more biologically relevant results compared to the baseline, allowing biologists to interpret the states as bindings to different protein families. As an example of non-medical applications, \cite{kdd} mentions the interpretability of a personalized HMM, i.e., an HMM with an extra hidden sequence layer revealing different transition regularities for each user, where the entropy of the emission probabilities between the two hidden sequences is used as a measure of interpretability. 

In the previous studies, the interpretability of HMMs is never defined and rarely measured, and in case the latter is done, the metric is typically domain-specific or adapted to a specific architecture of the graphical model. Hence, there is a lack of metrics for the interpretability of generic HMMs (or MHMMs). Without a proper way of measuring interpretability, the evaluation of novel contributions becomes difficult and it becomes practically impossible to devise techniques that directly optimize this property. In this work, we focus on interpretability of an MHMM, concerning a specific component of it, namely the state transition probability matrices. We argue that the entropy of these matrices is connected to interpretability; low entropy corresponds to that higher weights have been assigned to certain transitions, which hence can describe typical patterns among the sequences in the corresponding cluster, while high entropy corresponds to more uniformly weighted transitions, providing little information about the underlying patterns. Building on this idea, our first contribution is the proposal of an information-theoretic measure w.r.t state transitions for interpretability, similar to what was proposed by \cite{jordan}, however, w.r.t. feature selection in the model. Here, state transition importance is used instead of feature importance, so that the interpretability not only involves the important states but also their relations (transitions). To measure the state transition importance, we use the entropy (the average level of information inherent in the outcome of the model parameters), similar to the aforementioned work by \cite{kdd}, of Markov chains which involve state transitions; low entropic state transitions implies containing high weighted state transitions, indicating high state transition importance. Equivalently, low entropy implies high \textit{information} \cite{information}---contributing in interpretability by reducing the ``surprise'' incurred by clustering. Our second contribution is a regularized EM, the Informational Mixture of Hidden Markov Models (iMHMM), which reduces the entropy measure during the EM; i.e., the new EM assigns higher weight or importance (than what the standard EM assigns) on commonly transitioned states during the EM. Our approach is well-aligned with the idea of developing richer objective functions, in this case within EM, to provide built-in support for interpretable algorithms \cite{inter-survey}.

We compare our iMHMM to the standard EM for MHMMs and the recently developed Orthogonal Mixture of Hidden Markov Models (oMHMM) \cite{oMHMM}, as, to the best of our knowledge, no other approaches to infer MHMMs (merely using EM framework) have been introduced in the literature. The oMHMM is a regularized EM which encourages to increase the orthogonality of (distance between) the transition matrices of each pair of clusters. Even though the iMHMM is explicitly targeting to decrease the entropy of the clusters compared to the baseline, the oMHMM could potentially also decrease the entropy. It is expected that, as a by-product, the orthogonality penalty in the oMHMM can actually push the probabilities in the transition matrices away from entropic values. 

In the next section, we provide the notation and background of HMMs and MHMMs. We then introduce the novel measure of interpretability for MHMMs and the novel approach. Subsequently, we discuss the entropy of Markov chains in the oMHMM. In the Experiment section, we evaluate and compare the iMHMM to the two baselines. Finally, in the last section, we summarize the main findings and point out directions for future research.

\section{Preliminaries}
\subsection{Hidden Markov Models}
A finite Hidden Markov Model \cite{hmm} is a probabilistic graphical model which is formed by a sequence of observed symbols (observation sequence denoted as $Y = \{y_1,\dots,y_M\}$) and a sequence of hidden states (state sequence denoted as $C = \{c_1,\dots,c_M\}$) emitting $Y$, where $M$ is the length of the sequences. The hidden states follow a Markov structure, i.e., the occurrence of each state depends only on the state attained in the previous event. Each element (state) in $C$ takes a value, $j$,  which is between 1 and $J$. We denote each state at position m with value j as $c_{m,j}$, for $m = 1, \ldots, M$. The HMM is parameterized by $\rho$, $A$, and $O$ (initial, transition, and emission probabilities) which are denoted as $p(c_{1,j})$, $p(c_{m,j}|c_{m-1 ,i})$, and $p(y_m|c_{m,j})$, respectively. We refer jointly to these parameters as $\theta$. Maximum likelihood estimation \cite{bishop} is a common approach to infer the model parameters and it can be realized by maximizing the likelihood's lower bound using the Expectation Maximization (EM) algorithm \cite{bishop}. In EM, two phases are performed iteratively until convergence. The first phase, the \textit{E-step}, is to calculate the lower bound $Q(\theta, \theta^{old}) =  E_{C | Y, \theta^{old}}\big[\log P(Y, C |\theta)\big]$, the expected value of $\log P(Y, C |\theta)$ w.r.t. the conditional distribution of the hidden states given the observed data and old parameter estimates ($\theta^{old}$). The second phase, the \textit{M-step}, is to maximize the lower bound, i.e., the $Q$ function calculated in the $E$-step, w.r.t. $\theta$. Here, we focus on the maximization problem regarding the transition probabilities, $A$, as the proposed method involves parameter $A$. Therefore, the maximization problem shown in Eq. \ref{ll-a-hmm} results from maximizing the $Q$ function w.r.t. $A_{ij}$ which is the transition probability of moving from state $i$ to state $j$; the optimization is subject to the constraint $\sum^J_{j=1} A_{ij} = 1 $. Note that $\varepsilon(c_{m-1 ,i},c_{m,j})$ is the joint posterior distribution of two successive latent variables, calculated in the $E$-step; for further details, see \cite{bishop}.
\begin{equation}
    \max_{A_{ij}} \quad \sum^M_{m=2} \sum^J_{i=1} \sum^J_{j=1} \varepsilon(c_{m-1 ,i},c_{m,j}) \log A_{ij}
\label{ll-a-hmm}
\end{equation}

\subsection{Mixtures of Hidden Markov Models}

A Mixture of Hidden Markov Models (MHMM) is a probabilistic graphical model comprising a set of observation sequences, where each observation sequence (denoted as $Y_n = \{y_{n1}, \ldots, y_{nM} \}$) is generated by a \textit{mixture model} \cite{bishop}. The model contains $N$ observation sequences and $K$ components, where each component represents a cluster, defined by a unique set of HMM parameters. We define $Z_n$ as the latent variable concerning the cluster assignment of $Y_n$. Each $Y_n$, emitted from the state sequence $C_n = \{c_{1}^n, \ldots, c_{M}^n \}$, belongs to the $k$-th ($k$ takes a value between 1 and $K$) component and stems from an HMM parameterized by $\rho_k$, $O_k$, $A_k$. Finally, the probability of the observation sequences belonging to component $k$, the mixing coefficient, is denoted as $\pi_k$ with $\sum_{k=1}^K \pi_k =1$. To infer the parameters of the MHMM ($\rho_{1:K}$, $O_{1:K}$, $A_{1:K}$, $\pi_{1:K}$), EM can be performed similar to the previous section. As the $E$-step concerns calculating $Q(\theta, \theta^{old})$, we derive $E_{C, Z | Y, \theta^{old}}\big[\log p(Y, C, Z | \theta)\big]$. In the $M$-step, the $Q$ function is maximized. As in the previous section, we focus on the $M$-step in which the maximization problem concerns the transition matrices, i.e., maximizing the $Q$ function w.r.t. $A_{kij}$ which is the probability of moving from state $i$ to $j$ in component $k$. The maximization is formulated in Eq. \ref{ll-a-mhmm} subject to $\sum^J_{j=1} A_{kij} = 1 $.
\begin{equation}
    \max_{A_{kij}} \quad \sum^K_{k=1} \sum^N_{n=1} \sum^M_{m=2} \sum^J_{i=1} \sum^J_{j=1} \varepsilon_k(c_{m-1 ,i}^{n},c_{m,j}^{n}) \log A_{kij}
\label{ll-a-mhmm}
\end{equation}

\noindent Eq. \ref{ll-a-mhmm} has a closed-form solution which is the following:
\begin{equation}
    A_{kij} = \frac{\sum^N_{n=1} \sum^M_{m=2} \varepsilon_k(c_{m-1 ,i}^{n},c_{m,j}^{n})}{\sum^J_{j=1} \sum^N_{n=1} \sum^M_{m=2} \varepsilon_k(c_{m-1 ,i}^{n},c_{m,j}^{n})}
\label{a-update}
\end{equation}
For details on the rest of the update equations of the parameters during EM, we refer to \cite{spaMHMM}. Notice that the posterior probability of each observation sequence belonging to component $k$, $p(Z_n = k | Y_n, \theta^{old})$, can be calculated using the model parameters, $\theta^{old}$. The overview of the EM algorithm is shown in Algorithm \ref{em0}.

\begin{algorithm}
\small
\caption{EM for MHMM}
\begin{algorithmic}[1]
\Procedure{learn}{$Y_{1:N}$}:
\\Initialise $\theta$ := \{$\rho$, $O$, $A$, $\pi$\}
\Repeat
\State E-step: calculate $E_{C, Z | Y_{1:N}, \theta^{old}}\big[\log p(Y_{1:N}, C, Z | \theta)\big]$
\State M-step: update $\theta$ ($A$ calculated by Eq. \ref{a-update})
\Until {convergence}
\\ \Return \ $\theta$
\EndProcedure
\end{algorithmic}
\label{em0}
\end{algorithm}

\section{Interpretability of MHMMs}
We propose to use entropy of Markov chain as a measure for interpretability of MHMMs. The use of Markov chain is a natural choice because when clustering using EM for an MHMM, each cluster corresponds to a Markov chain which comprises a transition matrix, capturing underlying state transition patterns in data. Even though a Markov chain is parameterized by transition matrix $A$ and initial probabilities $\rho$, the entropy of a Markov chain is purely governed by $A$ \cite{information}. Calculation of the entropy of Markov chain depends on the assumption of the \textit{stationary} Markov chain---$A$ is not changed after training a sufficiently long sequence---which is often presumed when referring to HMMs and MHMMs. The stationary distribution is denoted as $\mu$ and is the left eigenvector of $A$ with an eigenvalue equal to one as formulated in Eq. \ref{stationary} \cite{information}. Each element of vector $\mu$ is the probability of being in a state $i$. The stationary distribution is achieved by solving Eq. \ref{stationary}.
\setlength{\abovedisplayskip}{2pt}
\setlength{\belowdisplayskip}{2pt}
\begin{equation}
    \mu^{T} = \mu^{T} A \iff \mu_i = \sum^J_{j=1} \mu_j A_{ij}, \quad \forall i
\label{stationary}
\end{equation}
According to \cite{information}, the entropy rate for a stationary Markov chain $C$ with transition matrix $A$ and stationary distribution $\mu$ is formulated in Eq. \ref{entropy}.
\setlength{\abovedisplayskip}{2pt}
\setlength{\belowdisplayskip}{2pt}
\begin{equation}
    H(C) = - \sum^J_{i=1} \mu_i \sum^J_{j=1} A_{ij} \log A_{ij}
\label{entropy}
\end{equation}
In this paper, we use $H(C)$ as the measure of entropy for each Markov chain in an MHMM. Low entropy implies stable states in the MHMM, hence interpretable state transition patterns. We now explain the interpretability of MHMM through an example. For a dataset being clustered by an MHMM, we calculate the clustering interpretability by $H(C)$ w.r.t. each component in MHMM. As each $H(C)$ calculates entropy w.r.t. a transition matrix, two matrices N and M (shown below) reveal the underlying transition patterns of the two clusters. Matrix N is interpretable, with near-deterministic values revealing self-transition (transition to the same state as the current one) patterns of the states (marked in bold), but, M due to high uncertainty reveals very little information about the state transition pattern. It can be concluded that one cluster can be interpreted as preserving states, indicated by the important state transitions being associated with high probability. The other cluster corresponds to entropic values (low importance or weight of state transitions) in M revealing no specific underlying pattern, hence, considered to be hard to interpret. These observations are confirmed by comparing the $H(C)$ calculated for the two clusters, where M results in high entropy (100\%), and N in low entropy (4\%). We define the entropy of the MHMM to be the average of the two entropy values (53\%). On average, this example of MHMM possesses interpretability to some extent but it is not fully interpretable. Namely, the important state transitions are the probabilities in bold, however for only one cluster.
\begin{equation*}
\quad
\quad
\quad
\quad
\quad
\quad
N =  
\begin{pmatrix}
\textbf{.99} & 0 & .01 \\
.1 & \textbf{.8} & .1 \\
.01 & 0 & \textbf{.99}
\end{pmatrix}
\quad
M  = 
\begin{pmatrix}
.5 & .2 & .3 \\
.4 & .3 & .3 \\
.2 & .4 & .4
\end{pmatrix}
\end{equation*}

\section{Regularized EM}

Given an MHMM, we seek to increase the interpretability by reducing the entropy of the model during clustering. This task involves updating transition matrices $A_{1:K}$ in EM; each transition matrix allows for interpreting the sequences belonging to the same cluster due to a common variation pattern. Concretely, the transition matrix summarizes the data points transition regularities throughout the sequences belonging to one cluster. Since transition to different states (in $A$) with similar probabilities imply high entropy, we need to push the probabilities towards values that are as dissimilar as possible. The natural way to proceed with this is to regularize the maximization problem in Eq. \ref{ll-a-mhmm} by subtracting the entropy (since we need to minimize the entropy), $H(C)$ in Eq. \ref{entropy}. The resulting problem corresponds to minimizing an objective function (simplified w.r.t. $A$) of form $- \log A -  \lambda A \log A$ where $\lambda > 0$ is the penalty hyperparameter; there can exist a $\lambda$, where the second derivative of the objective function can result in a matrix which is not positive semidefinite, i.e, non-convex optimization problem. For example, for $A_{11}=.5$ and $\lambda=10$, the second derivative w.r.t. $A_{11}$, $\frac{1}{A_{11}
^2} - \frac{\lambda}{A_{11}}$, is -16, i.e, a non-positive. A non-convex optimization problem cannot be solved by the standard EM due to the nature of the $Q$ function, i.e., convex optimization problem. Instead of solving the non-convex optimization problem, we propose a Maximum A Posteriori (MAP)---equivalent to regularization \cite{bishop,murphy}---approach, which results in a convex optimization problem. We refer to our regularized EM algorithm as the Informational Mixture of Hidden Markov Models (iMHMM). We regularize the objective function in Eq. \ref{ll-a-mhmm} by introducing a Dirichlet prior---with parameters jointly called $\eta$---on the transition matrix, $A$. The resulting optimization problem is shown in Eq. \ref{ll-a-mhmm-map}.
\begin{equation}
\begin{split}
    &\max_{A_{kij}} \quad \sum^K_{k=1} \sum^N_{n=1} \sum^M_{m=2} \sum^J_{i=1} \sum^J_{j=1} \textbf{A} + \textbf{D} \quad with: \\ &\textbf{A} = \varepsilon_k(c_{m-1 ,i}^{n},c_{m,j}^{n}) \log A_{kij} \\ &\textbf{D} =  \sum^K_{k=1} \sum^J_{i=1} \sum^J_{j=1} (\eta_{kij} - 1) \log A_{kij}
\end{split}
\label{ll-a-mhmm-map}
\end{equation}
By solving Eq. \ref{ll-a-mhmm-map}, the update equation for $A$ takes the form in Eq. \ref{map} \cite{ihmm}. Note that Eq. \ref{map} differs from Eq. \ref{a-update} by the parts involving the prior.
\begin{eqnarray}
\begin{split}
    &A_{kij} = \\
    &\frac{ \eta_{kij} - 1 + \sum^N_{n=1} \sum^M_{m=2} \varepsilon_k(c_{m-1 ,i}^{n},c_{m,j}^{n})}{ \sum^J_{j=1}(\eta_{kij} - 1) + \sum^J_{j=1} \sum^N_{n=1} \sum^M_{m=2} \varepsilon_k(c_{m-1 ,i}^{n},c_{m,j}^{n})}
\end{split}
\label{map}
\end{eqnarray}
As we intend to encapsulate our belief (decreasing entropy which can be achieved by having differently weighted state transitions rather than similarly weighted transitions) into a prior, we formulate the prior as the inclination towards the most common transition from each state, i.e., assigning weights on those transitions. Concretely, we reduce $\eta$ in Eq. \ref{map} to a single parameter $\lambda$, governing the amplification of our belief that depends on the the posterior probability $\varepsilon_k(c_{m-1 ,i}^{n},c_{m,j}^{n})$ revealing the most common transitions. We therefore define $\eta_{kij} = \lambda >= 1$ when the posterior probability $\varepsilon_k(c_{m-1 ,i}^{n},c_{m,j}^{n})$ is the maximum of its alternative values for $j = 1, \dots, J$; otherwise, $\eta_{kij} = 1$. This if-statement is denoted as the indicator function $\mathds{1}(\cdot)$ in the new update equation for $A$, taking the form in Eq. \ref{final-map}. Note that the reduction of multiple parameters in Eq. \ref{map} to a single parameter is done similar to \cite{ihmm}.
\begin{eqnarray}
\begin{split}
    &\gamma_j = \sum^N_{n=1} \sum^M_{m=2} \varepsilon_k(c_{m-1 ,i}^{n},c_{m,j}^{n}), \quad
    f = \mathds{1} \big (\gamma_j = \max_{l \in J}(\gamma_l))\big) \\ 
    &A_{kij} = \frac{ (\lambda - 1) f + \sum^N_{n=1} \sum^M_{m=2} \varepsilon_k(c_{m-1 ,i}^{n},c_{m,j}^{n})}{ (\lambda - 1) + \sum^J_{j=1} \sum^N_{n=1} \sum^M_{m=2} \varepsilon_k(c_{m-1 ,i}^{n},c_{m,j}^{n})}
\end{split}
\label{final-map}
\end{eqnarray}
We increase the probabilities of the commonly occurred transitions during training, by decreasing the probabilities of the uncommon transitions. This is achieved by applying a higher weight on the denominator of $A_{kij}$ when $f$ is zero, i.e., when the current state transition from $i$ to $j$ is not the most common or maximum transition in comparison with all transitions from $i$. In the proposed EM algorithm, Algorithm \ref{em-map}, we update the transition matrix according to the new update equation for $A$, however, only if the entropy of the Markov chain is reduced; otherwise, the original update equation is used. As the prior controls the entropy of each row of $A$ and not the whole $A$, we ensure the entropy reduction of the Markov chain (involving all of the elements of $A$) by the if-statement.
\begin{algorithm}
\small
\caption{iMHMM}
\begin{algorithmic}[1]
\Procedure{learn}{$Y_{1:N}$}:
\\Initialise $\theta$ := \{$\rho$, $O$, $A$, $\pi$\}
\Repeat
      \State E-step: calculate $E_{C, Z | Y_{1:N}, \theta^{old}}\big[\log p(Y_{1:N}, C, Z | \theta)\big]$
\State M-step: update $\theta$ (for updating $A$ perform the below) \State calculate $A_1$ using Eq. \ref{final-map} 
\State calculate entropy $H_1$ w.r.t $A_1$ using Eq. \ref{stationary} and \ref{entropy}
\State calculate $A_2$ using Eq. \ref{a-update}
\State calculate entropy $H_2$ w.r.t $A_2$ using Eq. \ref{stationary} and \ref{entropy}
      \If{$H_1 < H_2$}
          \State $A := A_1$
      \Else
          \State $A := A_2$
      \EndIf
\Until{convergence}
\\ \Return \ $\theta$
\EndProcedure
\end{algorithmic}
\label{em-map}
\end{algorithm}

\section{Interpretability of oMHMMs}
In this section, we explain, by an example, how a recent approach, called the Orthogonal Mixture of HMMs (oMHMM) \cite{oMHMM}, which was proposed for avoiding local optima by increasing the orthogonality of transition matrices, also may decrease the entropy of the Markov chains. Note that according to \cite{matrix}, the orthogonality of transition matrices is highest when their inner product, calculated by summing the elements of the matrix which is resulted by element-wise multiplication of the transition matrices, is zero. For a dataset (not shown here), transition matrices corresponding to two clusters are inferred. The standard EM for MHMM estimates matrices A and B (shown below); the oMHMM estimates A as in the standard EM, but C is estimated instead of B. After calculating the inner product and entropy, we see that A and C are more orthogonal and have a lower entropy than A and B. The inner product of A and C is near zero ($.99 \times .01 + .01 \times .99 + .02 \times .99 + .98 \times .01 = .04$) and their average entropy, calculated by Eq. \ref{stationary} and \ref{entropy}, is $.04$; in contrast, matrices A and B are less orthogonal, with an inner product of $.9$, and a higher average entropy of $.36$. It can be seen that the orthogonality penalty pushes the probabilities of C towards values of zero and one so that the sum of the element-wise multiplication of A and C can incline towards zero.
\begin{equation*}
\quad
\quad
\quad
\quad
A =  
\begin{pmatrix}
.99 & .01  \\
.02 & .98
\end{pmatrix}
\quad
B  = 
 \begin{pmatrix}
.4 & .6  \\
.5 & .5
\end{pmatrix}
\quad
C  = 
 \begin{pmatrix}
.01 & .99  \\
.99 & .01
\end{pmatrix}
\end{equation*}

\section{Experiments}

\subsection{Experimental Protocol}

We investigate the relative performance of the standard EM for MHMM (here referred to as MHMM), oMHMM, and iMHMM on real-world datasets \footnote[1]{For the implementation of iMHMM and the test results, we refer to \url{https://github.com/negar7918/iMHMM}}. 
Interpretability is measured by Markov chain entropy, while clustering performance is measured using the v-measure \cite{v-measure}. The v-measure quantifies to which extent a cluster only contains data points that are members of a single class (the higher v-measure the better clustering). Having access to the cluster labels, we use v-measure in favor of accuracy, since the former penalizes the performance for each wrong estimated cluster membership, while accuracy is mainly used for supervised learning and not clustering. Note that improving upon the entropy (the lower the entropy, the higher the interpretability) is meaningful when the clustering performance is not drastically reduced, i.e., failing in satisfying the primary goal, clustering. We also report the computation cost, measured by the number of EM iterations. 

For the penalty hyperparameters, we set $\lambda$ to one for oMHMM (as chosen in \cite{oMHMM}), and, to the length of the observation sequence---inspired by \cite{ihmm} for iMHMM. For the EM initializations of transition matrices, we use Dirichlet distribution prior with parameters set to 0.1 for each row of the transition matrix, to enable extreme probabilities which can, in general, be beneficial for reducing entropy of the corresponding Markov chain. In the table of results, we use the term entropy by which we mean the average entropy of the Markov chains within an MHMM; each Markov chain entropy is calculated using Eq. \ref{stationary} and \ref{entropy}. Finally, the result of the better performing model is highlighted in bold.

\subsection{Experiments with Biological Data}
\label{bio}

We perform clustering of cancer cells on a previously published biological dataset\footnote[2]{Available from the NCBI Sequence Read Archive under accession number SRP074289 \cite{navin}.}. The cells comprise two main clusters: primary colon tumor cells and metastatic cells migrated to the liver. Each cluster contains 18 cells. Moreover, the metastatic cluster has shown to comprise two sub-clusters \cite{navin}. We consider the primary, metastatic, and the sub-clusters for our clustering on CRC2 patient data using the genomic sequence of chromosome 4 and the sequence of chromosome 18 to 21. We refer to the first dataset as ``chrom 4'' and the second as ``chroms 18 to 21'' with 808 and 903 genomic regions (sequence length), respectively. Each region bears the characteristic of that region by a count number, that is, the data contains sequences of count numbers. For the purpose of clustering, we use a mixture of HMMs, where each cell sequence represents the observation sequence of an HMM. For the datasets ``chrom 4'' and ``chroms 18 to 21'', the number of hidden states of each HMM is three and six, respectively; the number of hidden states corresponds to the hidden underlying \textit{copy number} of chromosomes in \cite{navin}.

\begin{table}
\caption{Clustering of cancer cells using chromosomes.}
\centering
\footnotesize
\scalebox{0.8}
{
\begin{tabular}{ | c | c | c | c |}
 \multicolumn{4}{c}{\% V-measure}  \\ 
  \hline 
  dataset & MHMM & oMHMM & iMHMM \\
  \hline 
   chrom 4 & 4\% & \textbf{59}\% & \textbf{59}\%\\
   \hline 
   chroms 18 to 21 & 59\% & \textbf{100}\% & \textbf{100}\%\\
   \hline 
   *chroms 18 to 21 & 0\% & \textbf{67}\% & \textbf{67}\%\\
  \hline 
\end{tabular}}
\quad
\scalebox{0.8}
{
\begin{tabular}{ | c | c | c | c |}
 \multicolumn{4}{c}{\% Entropy}  \\ 
  \hline 
  dataset & MHMM & oMHMM & iMHMM \\
  \hline 
   chrom 4 & \textbf{10}\% & 13\% & 12\%\\
   \hline 
   chroms 18 to 21 & 47\% & 12\% & \textbf{8}\%\\
   \hline 
   *chroms 18 to 21 & 51\% & 28\% & \textbf{9}\%\\
  \hline 
\end{tabular}}
\quad
\scalebox{0.8}
{
\begin{tabular}{ | c | c | c | c |}
 \multicolumn{4}{c}{\# Iterations}  \\ 
  \hline 
  dataset & MHMM & oMHMM & iMHMM \\
  \hline 
   chrom 4 & 5 & \textbf{4} & \textbf{4}\\
   \hline 
   chrom 18 to 21 & 12 & \textbf{3} & \textbf{3}\\
   \hline 
   *chrom 18 to 21 & 4 & \textbf{3} & \textbf{3}\\
  \hline 
\end{tabular}}
\label{navin}
\end{table}
After performing the clustering of metastatic- and primary cells, using MHMM, oMHMM, and iMHMM, we calculate the resulting v-measure, entropy, and the number of EM iterations for each method. As shown in Table \ref{navin}, the two main clusters are detected using datasets ``chrom 4'' and ``chroms 18 to 21''; moreover, the methods are tested to detect the sub-clusters using ``chroms 18 to 21'' (the dataset is marked with a `*' to indicate the clustering of three groups instead of the two main ones). We can observe that iMHMM outperforms MHMM. Note that MHMM has a lower entropy for dataset ``chrom 4'', however, the corresponding clustering result is 4\% which is 55 percentage points less than that produced by iMHMM. Note that it is not very meaningful to consider the interpretability of a very poor clustering. In the experiments, iMHMM outperforms oMHMM w.r.t. entropy, while they perform on a similar level w.r.t. the v-measure and the number of iterations. The maximum improvement achieved by iMHMM in comparison to MHMM is 42 percentage points w.r.t. entropy and 67 percentage points w.r.t. v-measure. The low-entropy estimated transitions by iMHMM comply with the low-entropy heatmap of the corresponding calculated copy numbers in \ref{navin}. Due to the structural (non-entropic) nature of copy number variations, which can enable interpretability by inferring the underlying structure, iMHMM is hence suited in this respect for this type of data. Finally, the fewer number of iterations required by oMHMM and iMHMM, in comparison to those of MHMM, indicates a faster convergence of EM when using regularization. For the dataset ``chroms 18 to 21'', the number of iterations for iMHMM and oMHMM is reduced to 25\% of that for MHMM. In the above experiments, iMHMM compares favorably to the other methods.

\subsection{Experiments with Hand Movement Data}
We repeatedly cluster datasets obtained by restricting the ``Libras movement'' dataset with 15 classes of hand movements, in the UCI machine learning repository \cite{digit-data}, to two digits at a time, inspired by \cite{momMHMM}. We repeat the experiments for three numbers of hidden states (S = 2, 3, 4) in the HMMs to assess the performance by increasing S.
\begin{table}
\caption{Clustering of hand movements for 2 hidden states.}
\centering
\footnotesize
\scalebox{.8}
{
\begin{tabular}{ | c | c  c  c |c  c  c |}
 \multicolumn{7}{c}{}  \\ 
 \hline 
   & & \% V-measure & & & \% Entropy &\\
  \hline 
  dataset & MHMM & oMHMM & iMHMM  & MHMM & oMHMM & iMHMM \\
  \hline 
   1 vs 8 & 6\% & 22\% & \textbf{35}\% & 8\% & \textbf{4}\% & 5\%\\
   \hline 
   2 vs 3 & \textbf{3}\% & 1\% & \textbf{3}\% & 6\% & \textbf{4}\% & 5\%\\
   \hline 
   3 vs 4 & 5\% & 5\% & 5\% & 8\% & \textbf{2}\% & 5\%\\
   \hline 
   3 vs 8 & 1\% & 7\% & \textbf{13}\% & \textbf{0}\% & 1\% & 1\%\\
   \hline 
   4 vs 14 & 2\% & 2\% & 2\% & 3\% & \textbf{0}\% & 1\%\\
  \hline 
\end{tabular}}
\quad
\scalebox{.8}
{
\begin{tabular}{ | c | c | c | c |}
 \multicolumn{4}{c}{\# Iterations}  \\ 
  \hline 
  dataset & MHMM & oMHMM & iMHMM \\
  \hline 
   1 vs 8 & 6 & \textbf{2} & 3\\
   \hline 
   2 vs 3 & 4 & 4 & \textbf{3}\\
   \hline 
   3 vs 4 & 4 & \textbf{3} & 5\\
   \hline 
   3 vs 8 & 5 & \textbf{2} & 3\\
   \hline 
   4 vs 14 & \textbf{3} & \textbf{3} & 4\\
  \hline 
\end{tabular}}
\label{hand-2}
\end{table}
First, we compare MHMM, oMHMM and iMHMM using ``S = 2'', see Table \ref{hand-2}. We observe that iMHMM outperforms MHMM w.r.t. entropy---iMHMM performs better or equally good as MHMM w.r.t. the v-measure---for all datasets; for the ``3 vs 8'' dataset, MHMM almost fails to produce meaningful clusters (the v-measure is 1\%), hence, as aforementioned, it is not meaningful to consider the entropy and number of iterations of MHMM. Moreover, iMHMM performs better or equally good as oMHMM w.r.t. v-measure for all datasets, but, oMHMM performs slightly better than iMHMM w.r.t. entropy, which however may not outweigh the fact that iMHMM has a generally better clustering performance. The numbers of iterations required by all methods are similar, but with a slight advantage for oMHMM. 

Second, we compare the methods using ``S = 3'', see Table \ref{hand-3}. We observe that iMHMM and oMHMM outperform MHMM (but not each other) w.r.t. v-measure. Regarding entropy, iMHMM outperforms MHMM for the ``1 vs 8'' dataset; for the remaining datasets, the entropies reported for MHMM are left without consideration due to the corresponding 0\% and 1\% v-measures. The entropies resulting from oMHMM and iMHMM are similar but we can observe that iMHMM is performing slightly better. The numbers of iterations for the three methods are close to each other but, in general, oMHMM and iMHMM outperform MHMM.

Finally, we compare the methods using ``S = 4'', see Table \ref{hand-4}. We observe that iMHMM outperforms MHMM w.r.t. v-measure and entropy for the ``1 vs 8'', ``2 vs 3'', and ``4 vs 14'' datasets. All methods fail to cluster the ``3 vs 8'' dataset; hence, the results for the other measures are left without consideration. For the ``3 vs 4'' dataset, MHMM outperforms iMHMM w.r.t. v-measure---an example when iMHMM is not beneficial. However, for this dataset, oMHMM performs clustering as good as MHMM and even outperforms it w.r.t. entropy. Aside from this dataset, iMHMM performs better or equal to oMHMM w.r.t. v-measure and outperforms oMHMM w.r.t. entropy. The number of iterations among all methods are close to each other, but for the ``3 vs 4'' dataset, oMHMM consumes twice more iterations.
\begin{table}
\caption{Clustering of hand movements 3 hidden states.}
\centering
\footnotesize
\scalebox{.8}
{
\begin{tabular}{ | c | c  c  c |c  c  c |}
 \multicolumn{7}{c}{}  \\ 
 \hline 
   & & \% V-measure & & & \% Entropy &\\
  \hline 
  dataset & MHMM & oMHMM & iMHMM  & MHMM & oMHMM & iMHMM \\
  \hline  
   1 vs 8 & 6\% & \textbf{13}\% & \textbf{13}\% & 10\% & 7\% & \textbf{6}\%\\
   \hline 
   2 vs 3 & 0\% & \textbf{13}\% & \textbf{13}\% & \textbf{9}\% & 10\% & \textbf{9}\%\\
   \hline 
   3 vs 4 & 1\% & \textbf{23}\% & \textbf{23}\% & 6\% & 6\% & \textbf{5}\%\\
   \hline 
   3 vs 8 & 0\% & \textbf{15}\% & \textbf{15}\% & \textbf{0}\% & 4\% & 3\%\\
   \hline 
   4 vs 14 & 1\% & \textbf{15}\% & \textbf{15}\% & 1\% & 1\% & 1\%\\
  \hline 
\end{tabular}}
\quad
\scalebox{.8}
{
\begin{tabular}{ | c | c | c | c |}
 \multicolumn{4}{c}{\# Iterations}  \\ 
  \hline 
  dataset & MHMM & oMHMM & iMHMM \\
  \hline 
   1 vs 8 & 3 & \textbf{2} & \textbf{2}\\
   \hline 
   2 vs 3 & 4 & \textbf{3} & \textbf{3}\\
   \hline 
   3 vs 4 & 4 & \textbf{2} & \textbf{2}\\
   \hline 
   3 vs 8 & 3 & \textbf{2} & \textbf{2}\\
   \hline 
   4 vs 14 & 3 & \textbf{2} & 3\\
  \hline 
\end{tabular}}
\label{hand-3}
\caption{Clustering of hand movements for 4 hidden states.}
\centering
\footnotesize
\scalebox{.8}
{
\begin{tabular}{ | c | c  c  c |c  c  c |}
 \multicolumn{7}{c}{}  \\ 
 \hline 
   & & \% V-measure & & & \% Entropy &\\
  \hline 
  dataset & MHMM & oMHMM & iMHMM  & MHMM & oMHMM & iMHMM \\
  \hline
   1 vs 8 & 2\% & 0\% & \textbf{8}\% & 8\% & 8\% & \textbf{7}\%\\
   \hline 
   2 vs 3 & 2\% & \textbf{4}\% & \textbf{4}\% & 19\% & 8\% & \textbf{6}\%\\
   \hline 
   3 vs 4 & \textbf{26}\% & \textbf{26}\% & 4\% & 10\% & 6\% & \textbf{3}\%\\
   \hline 
   3 vs 8 & 0\% & 0\% & 0\% & 12\% & 3\% & \textbf{1}\%\\
   \hline 
   4 vs 14 & 7\% & 0\% & \textbf{12}\% & 8\% & 3\% & \textbf{0}\%\\
  \hline 
\end{tabular}}
\quad
\scalebox{.8}
{
\begin{tabular}{ | c | c | c | c |}
 \multicolumn{4}{c}{\# Iterations}  \\ 
  \hline 
  dataset & MHMM & oMHMM & iMHMM \\
  \hline 
   1 vs 8 & \textbf{3} & 5 & 4\\
   \hline 
   2 vs 3 & 6 & \textbf{5} & \textbf{5}\\
   \hline 
   3 vs 4 & \textbf{3} & 7 & 4\\
   \hline 
   3 vs 8 & 5 & \textbf{4} & \textbf{4}\\
   \hline 
   4 vs 14 & 3 & 3 & 3\\
  \hline 
\end{tabular}}
\label{hand-4}
\end{table}

From the above results,
we can conclude that iMHMM and oMHMM in most cases outperform MHMM. But there are exceptions, such as for the ``3 vs 4'' dataset when assuming four hidden states, in which iMHMM does not improve upon MHMM. Note that one may remove the effect of regularization in iMHMM, and hence obtain the same performance as MHMM, by setting the penalty hyperparameter to zero, e.g. following hyperparameter tuning, which however has not been performed in this study. After plotting movement 3 and 4 (not shown here), an oscillating pattern in data is revealed. The oscillation pattern can cause the iMHMM to perform worse, since iMHMM pushes towards low entropy and assigns transition probabilities of 0 and 1 to the oscillating states which should instead posses entropic probabilities.  


\section{Concluding Remarks} 

We propose to use entropy of Markov chain as a proxy for interpretability of MHMMs. We investigate this measure when using EM and a previously optimized EM; moreover, we propose a novel regularized EM to improve the interpretability by reducing the entropy, which leads to assigning higher weights (importance) to commonly state transitions, than those in the standard EM. Empirically, we show that it is possible to improve the interpretability of MHMMs, as measured by entropy, without sacrificing (but rather improving) clustering performance and computational costs, measured by v-measure and number of EM iterations, respectively.



The use of entropy as a measure of interpretability and its relation to other measures should be investigated further, both qualitatively and quantitatively. In particular, the proposed regularized EM should be compared to the use of other possible metrics to capture the interpretability of MHMMs, e.g., considering the number of hidden states. Other directions for future work include solving the aforementioned non-convex regularization using disciplined convex-concave programming (DCCP), \cite{dccp}, and, improving \textit{identifiability}---the ability to distinguish the labels of the hidden variable with equal corresponding likelihoods---\cite{murphy}.

\bibstyle{aaai}
\bibliography{main}

\end{document}